# $AltAlt^p$: Online Parallelization of Plans with Heuristic State Search

**Romeo Sanchez Nigenda**                                    RSANCHEZ@ASU.EDU
**Subbarao Kambhampati**                                      RAO@ASU.EDU
*Department of Computer Science and Engineering,*
*Arizona State University, Tempe AZ 85287-5406*

## Abstract

Despite their near dominance, heuristic state search planners still lag behind disjunctive planners in the generation of parallel plans in classical planning. The reason is that directly searching for parallel solutions in state space planners would require the planners to branch on all possible subsets of parallel actions, thus increasing the branching factor exponentially. We present a variant of our heuristic state search planner *AltAlt* called $AltAlt^p$ which generates parallel plans by using greedy online parallelization of partial plans. The greedy approach is significantly informed by the use of novel distance heuristics that $AltAlt^p$ derives from a graphplan-style planning graph for the problem. While this approach is not guaranteed to provide optimal parallel plans, empirical results show that $AltAlt^p$ is capable of generating good quality parallel plans at a fraction of the cost incurred by the disjunctive planners.

## 1. Introduction

Heuristic state space search planning has proved to be one of the most efficient planning frameworks for solving large deterministic planning problems (Bonet, Loerincs, & Geffner, 1997; Bonet & Geffner, 1999; Bacchus, 2001). Despite its near dominance, its one achilles heel remains generation of "parallel plans" (Haslum & Geffner, 2000). Parallel plans allow concurrent execution of multiple actions in each time step. Such concurrency is likely to be more important as we progress to temporal domains. While disjunctive planners such as Graphplan (Blum & Furst, 1997) SATPLAN (Kautz & Selman, 1996) and GP-CSP (Do & Kambhampati, 2000) seem to have no trouble generating such parallel plans, planners that search in the space of states are overwhelmed by this task. The main reason is that straightforward methods for generation of parallel plans would involve progression or regression over sets of actions. This increases the branching factor of the search space exponentially. Given $n$ actions, the branching factor of a simple progression or regression search is bounded by $n$, while that of progression or regression search for parallel plans will be bounded by $2^n$.

The inability of state search planners in producing parallel plans has been noted in the literature previously. Past attempts to overcome this limitation have not been very successful. Indeed, Haslum and Geffner (2000) consider the problem of generating parallel plans using a regression search in the space of states. They note that the resulting planner, HSP*p, scales significantly worse than Graphplan. They present TP4 in (Haslum & Geffner, 2001), which in addition to being aimed at actions with durations, also improves the branching scheme of HSP*p, by making it incremental along the lines of Graphplan. Empirical studies reported by Haslum and Geffner (2001), however indicate that even this new approach, unfortunately, scales quite poorly compared to Graphplan variants. Informally, this achilles heel of heuristic state search planners has been interpreted as a sort of last stand of the disjunctive planners – *only they are capable of generating parallel plans efficiently.*

Given that the only way of efficiently generating optimal parallel plans involves using disjunctive planners, we might want to consider ways of generating near-optimal parallel plans using state search planners. One obvious approach is to post-process the sequential plans generated by the state search planners to make them parallel. While this can easily be done - using approaches such as those explored by Backstrom (1998), one drawback is that such approaches are limited to transforming the sequential plan given as input. Parallelization of sequential plans often results in plans that are not close to optimal parallel plans.[1]

An alternative, that we explore in this paper, involves incremental online parallelization. Specifically, our planner *AltAlt$^p$*, which is a variant of the *AltAlt* planner (Sanchez, Nguyen, & Kambhampati, 2000; Nguyen, Kambhampati, & Sanchez, 2002), starts its search in the space of regression over single actions. Once the most promising single action to regress is selected, *AltAlt$^p$* then attempts to parallelize ("fatten") the selected search branch with other independent actions. This parallelization is done in a greedy incremental fashion - actions are considered for addition to the current search branch based on the heuristic cost of the subgoals they promise to achieve. The parallelization continues to the next step only if the state resulting from the addition of the new action has a better heuristic cost. The sub-optimality introduced by the greedy nature of the parallelization is offset to some extent by a plan-compression procedure called *Pushup* that tries to rearrange the evolving parallel plans by pushing up actions to higher levels in the search branch (i.e. later stages of execution) in the plan.

Despite the seeming simplicity of our approach, it has proven to be quite robust in practice. In fact, our experimental comparison with five competing planners - STAN (Long & Fox, 1999), LPG (Gerevini & Serina, 2002), Blackbox (Kautz & Selman, 1996), SAPA (Do & Kambhampati, 2001) and TP4 (Haslum & Geffner, 2001) - shows that *AltAlt$^p$* is a viable and scalable alternative for generating parallel plans in several domains. For many problems, *AltAlt$^p$* is able to generate parallel plans that are close to optimal in makespan. It also seems to retain the efficiency advantages of heuristic state search over disjunctive planners, producing plans in a fraction of the time taken by the disjunctive planners in many cases. *AltAlt$^p$* has also been found to be superior to post-processing approaches. Specifically, we compared *AltAlt$^p$* to an approach that involves post-processing the sequential plans generated by *AltAlt* using the techniques from Backstrom (1998). We found that *AltAlt$^p$* is able to generate shorter parallel plans in many cases. Finally, we show that *AltAlt$^p$* incurs very little additional overhead compared to *AltAlt*.

In the rest of this paper, we discuss the implementation and evaluation of our approach to generate parallel plans with *AltAlt$^p$*. Section 2 starts by providing a review of the *AltAlt* planning system, on which *AltAlt$^p$* is based. Section 3 describes the generation of parallel plans in $AltAlt^p$. Section 4 presents extensive empirical evaluation of $AltAlt^p$. The evaluation includes both comparison and ablation studies. Finally, Section 5 discusses some related work in classical as well as metric temporal planning. Section 6 summarizes our contributions.

---

1. We will empirically demonstrate this later; curious readers may refer to the plots in Figure 15.

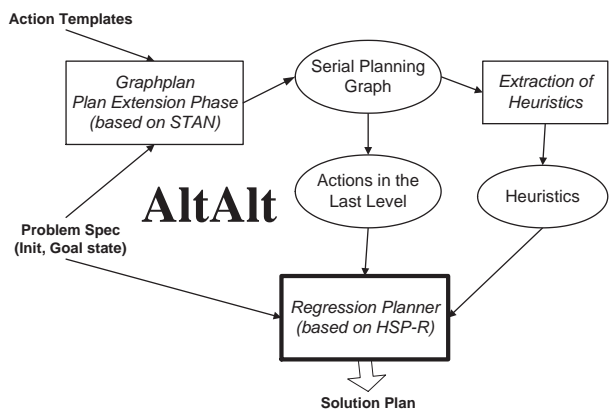

Figure 1: Architecture of *AltAlt*

## 2. *AltAlt* Background Architecture and Heuristics

The *AltAlt* planning system is based on a combination of Graphplan (Blum & Furst, 1997; Long & Fox, 1999; Kautz & Selman, 1999) and heuristic state space search (Bonet et al., 1997; Bonet & Geffner, 1999; McDermott, 1999) technology. *AltAlt* extracts powerful heuristics from a planning graph data structure to guide a regression search in the space of states. The high level architecture of *AltAlt* is shown in Figure 1. The problem specification and the action template description are first fed to a Graphplan-style planner (in our case, STAN from Long & Fox, 1999), which constructs a planning graph for that problem in polynomial time (we assume the reader is familiar with the Graphplan algorithm of Blum & Furst, 1997). This planning graph structure is then fed to a heuristic extractor module that is capable of extracting a variety of effective heuristics (Nguyen & Kambhampati, 2000; Nguyen et al., 2002). These heuristics, along with the problem specification, and the set of ground actions in the final action level of the planning graph structure are fed to a regression state-search planner.

To explain the operation of *AltAlt* at a more detailed level, we need to provide some further background on its various components. We shall start with the regression search module. Regression search is a process of searching in the space of potential plan suffixes. The suffixes are generated by starting with the goal state and regressing it over the set of relevant action instances from the domain. The resulting states are then (non-deterministically) regressed again over relevant action instances, and this process is repeated until we reach a state (set of subgoals) which is satisfied by the initial state. A state $S$ in our framework is a set of (conjunction of) literals that can be seen as "subgoals" that need to be made true on the way to achieving the top level goals. An action instance $a$ is considered relevant to a state $S$ if the effects of $a$ give at least one element of $S$ and do not delete *any element* of $S$. The result of regressing $S$ over $a$ is then $(S \backslash eff(a)) \cup prec(a)$ - which is essentially the set of goals that still need to be achieved before the application of $a$, such that everything in $S$ would have been achieved once $a$ is applied. For each relevant action $a$, a separate search branch is generated, with the result of regressing $S$ over that action as the new fringe in that branch. Search terminates with success at a node if every literal in the state corresponding to that node is present in the initial state of the problem.

The crux of controlling the regression search involves providing a heuristic function that can estimate the relative goodness of the states on the fringe of the current search tree and guide the search in the most promising directions. The heuristic function needs to evaluate the cost of achieving the set $S$ of subgoals (comprising a regressed state) from the initial state. In other words, the heuristic computes the length of the plan needed to achieve the subgoals from the initial state. We now discuss how such a heuristic can be computed from the planning graph, which, provides optimistic reachability estimates.

Normally, the planning graph data structure supports "parallel" plans - i.e., plans where at each step more than one action may be executed simultaneously. Since we want the planning graph to provide heuristics to the regression search module of *AltAlt*, which generates sequential solutions, we first make a modification to the algorithm so that it generates a "serial planning graph." A *serial planning graph* is a planning graph in which, in addition to the normal mutex relations, every pair of non-noop actions at the same level are marked mutex. These additional action mutexes propagate to give additional propositional mutexes. Finally, a planning graph is said to **level off** when there is no change in the action, proposition and mutex lists between two consecutive levels.

We will assume for now that given a problem, the Graphplan module of *AltAlt* is used to generate and expand a serial planning graph until it levels off. As discussed by Sanchez et al. (2000), we can relax the requirement of growing the planning graph to level-off, if we can tolerate a graded loss of informedness of heuristics derived from the planning graph. We will start with the notion of level of a set of propositions:

**Definition 1 (Level)** *Given a set $S$ of propositions, $lev(S)$ is the index of the first level in the leveled serial planning graph in which all propositions in $S$ appear and are non-mutex with one another. If $S$ is a singleton, then $lev(S)$ is just the index of the first level where the singleton element occurs. If no such level exists, then $lev(S) = \infty$ if the planning graph has been grown to level-off.*

The intuition behind this definition is that the level of a literal $p$ in the serial planning graph provides a lower bound on the length of the plan (which, for a serial planning graph, is equal to the number of actions in the plan) to achieve $p$ from the initial state. Using this insight, a simple way of estimating the cost of a set of subgoals will be to sum their levels.

**Heuristic 1 (Sum heuristic)** $h_{sum}(S) := \sum_{p \in S} lev(\{p\})$

The sum heuristic is very similar to the greedy regression heuristic used in UNPOP (McDermott, 1999) and the heuristic used in the HSP planner (Bonet et al., 1997). Its main limitation is that the heuristic makes the implicit assumption that all the subgoals (elements of $S$) are independent. The $h_{sum}$ heuristic is neither admissible nor particularly informed as it ignores the interactions between the subgoals. To develop more effective heuristics, we need to consider both positive and negative interactions among subgoals in a limited fashion.

In (Nguyen et al., 2002), we discuss a variety of ways of using the planning graph to incorporate negative and positive interactions into the heuristic estimate, and discuss their relative tradeoffs. One of the best heuristics according to that analysis was a heuristic called $h_{AdjSum2M}$. We adopted this heuristic as the default heuristic in *AltAlt*. The basic idea of $h_{AdjSum2M}$ is to adjust the sum heuristic to take positive and negative interactions into account. This heuristic approximates the cost of achieving the subgoals in some set $S$ as the sum of the cost of achieving $S$, while considering positive interactions and ignoring negative interactions, plus a penalty for ignoring the negative interactions. The first component $RP(S)$ can be computed as the length of a "relaxed plan" for supporting $S$, which is extracted by *ignoring all the mutex relations*. To approximate the penalty induced by the negative interactions alone, we proceed with the following argument. Consider any pair of subgoals $p, q \in S$. If there are no negative interactions between $p$ and $q$, then $lev(\{p, q\})$, the level at which $p$ and $q$ are present together, is exactly the maximum of $lev(p)$ and $lev(q)$. The degree of negative interaction between $p$ and $q$ can thus be quantified by:

$$\delta(p, q) = lev(\{p, q\}) - max\left(lev(p), lev(q)\right)$$

We now want to use the $\delta$ - values to characterize the amount of negative interactions present among the subgoals of a given set $S$. If all subgoals in $S$ are pair-wise independent, clearly, all $\delta$ values will be zero, otherwise each pair of subgoals in $S$ will have a different value. The largest such $\delta$ value among any pair of subgoals in $S$ is used as a measure of the negative interactions present in $S$ in the heuristic $h_{AdjSum2M}$. In summary, we have

**Heuristic 2 (Adjusted 2M)** $h_{AdjSum2M}(S) := length(RP(S)) + max_{p,q \in S}\delta(p, q)$

The analysis by Nguyen et al. (2002) shows that this is one of the more robust heuristics in terms of both solution time and quality. This is thus the default heuristic used in *AltAlt* (as well as *AltAlt$^p$*; see below).

## 3. Generation of Parallel Plans Using *AltAlt$^p$*

The obvious way to make *AltAlt* produce parallel plans would involve regressing over subsets of (non interfering) actions. Unfortunately, this increases the branching factor exponentially and is infeasible in practice. Instead, *AltAlt$^p$* uses a greedy depth-first approach that makes use of its heuristics to regress single actions, and incrementally parallelizes the partial plan at each step, rearranging the partial plan later if necessary.

The high level architecture of *AltAlt$^p$* is shown in Figure 2. Notice that the heuristic extraction phase of *AltAlt$^p$* is very similar to that of *AltAlt*, but with one important modification. In contrast to *AltAlt* which uses a "serial" planning graph as the basis for its heuristic (see Section 2), *AltAlt$^p$* uses the standard "parallel" planning graph. This makes sense given that *AltAlt$^p$* is interested in parallel plans while *AltAlt* was aimed at generating sequential plans. The regression state-search engine for *AltAlt$^p$* is also different from the search module in *AltAlt*. *AltAlt$^p$* augments the search engine of *AltAlt* with 1) a fattening step and 2) a plan compression procedure (*Pushup*). The details of these procedures are discussed below.

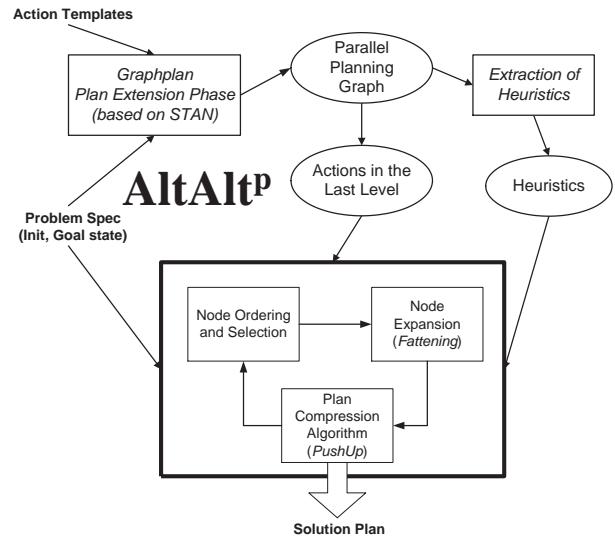

Figure 2: Architecture of *AltAlt*[p]

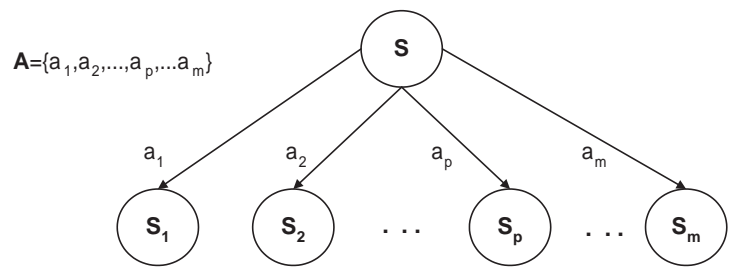

Figure 3: *AltAlt*[p] Notation

**parexpand**(S)
    A ← *get set of applicable actions for current state $S$*
    **forall** $a_i \in A$
        $S_i \leftarrow Regress(S, a_i)$
        CHILDREN(S) ← CHILDREN(S) + $S_i$
    $S_p \leftarrow$ *The state among $Children(S)$ with minimum*
    $h_{adjsum2M}$ *value*
    $a_p \leftarrow$ *the action that regresses to $S_p$ from $S$*
            /**Fattening process
    $O \leftarrow \{\ a_p\ \}$
    **forall** $g \in S$ ranked in the decreasing order of $level(g)$
            Find an action $a_g \in A$ supporting $g$ such that $a_g \notin O$ and
            $a_i$ is pairwise independent with each action in $O$.
            If there are multiple such actions, pick the one which has
            minimum $h_{adjsum}(Regress(S, O + a_g))$ among all $a_g \in A$
            If $h_{adjsum2M}(S, O + a_i) < h_{adjsum2M}(S, O)$
                $O \leftarrow O + a_g$
    $S_{par} \leftarrow Regress(S, O)$
    CHILDREN(S) ← CHILDREN(S) + $S_{par}$
    **return** CHILDREN
**END**;

Figure 4: Node Expansion Procedure

The general idea in *AltAlt$^p$* is to select a fringe action $a_p$ from among those actions $A$ used to regress a particular state $S$ during any stage of the search (see Figure 3). Then, the pivot branch given by the action $a_p$ is "fattened" by adding more actions from $A$, generating a new state that is a consequence of regression over multiple parallel actions. The candidate actions used for fattening the pivot branch must (a) come from the sibling branches of the pivot branch and (b) be pairwise independent with all the other actions currently in the pivot branch. We use the standard definition of action independence: two actions $a_1$ and $a_2$ are considered independent if the state $S'$ resulting after regressing both actions simultaneously is the same as that obtained by applying $a_1$ and $a_2$ sequentially with any of their possible linearizations. A sufficient condition for this is that the preconditions and effects of the actions do not interfere:

$$((|prec(a_1)| \cup |eff(a_1)|) \cap (|prec(a_2)| \cup |eff(a_2)|)) = \emptyset$$

where |L| refers to the non-negated versions of the literals in the set L. We now discuss the details of how the pivot branch is selected in the first place, and how the branch is incrementally fattened.

**Selecting the Pivot Branch:** Figure 4 shows the procedure used to select and parallelize the pivot branch. The procedure first identifies the set of regressable actions $A$ for the current node $S$, and regresses each of them, computing the new children states. Next, the action leading to the child state with the lowest heuristic cost among the new children is selected as the pivot action $a_p$, and the corresponding branch becomes the pivot branch.

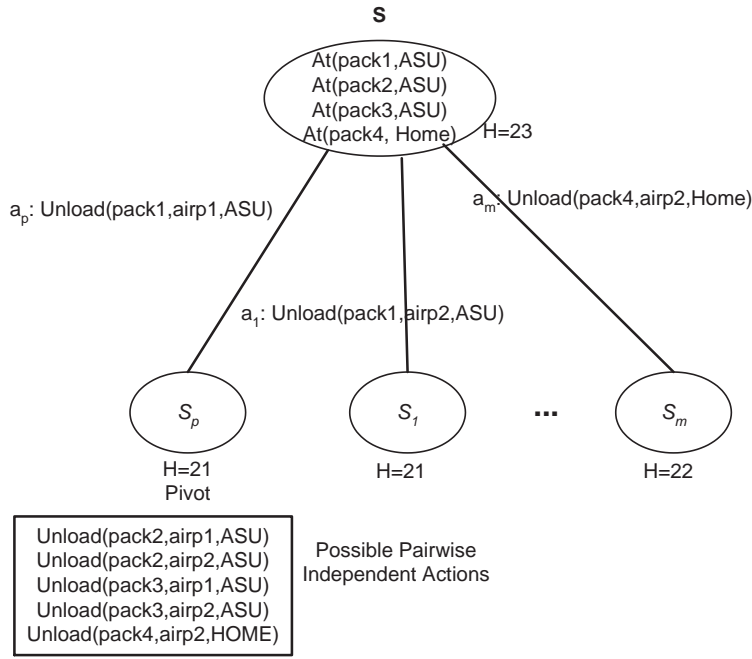

Figure 5: After the regression of a state, we can identify the *Pivot* and the related set of pairwise independent actions.

The heuristic cost of the states is computed with the $h_{adjsum2M}$ heuristic from *AltAlt*, based on a "parallel" planning graph. Specifically, in the context of the discussion of the $h_{adjsum2M}$ heuristic at the end of Section 2, we compute the $\delta(p, q)$ values, which in turn depend on the $level(p), level(q)$ and $level(p, q)$ in terms of the levels in the parallel planning graph rather than the serial planning graph. It is easy to show that the level of a set of conditions on the parallel planning graph will be less than or equal to the level on the serial planning graph. The length of the relaxed plan is still computed in terms of number of actions. We show later (see Figure 19(a)) that this change does improve the quality of the parallel plans produced by *AltAlt$^p$*.

The search algorithm used in *AltAlt$^p$* is similar to that used in HSPr (Bonet & Geffner, 1999) - it is a hybrid between greedy depth first and a weighted A* search. It goes depth-first as long as the heuristic cost of any of the children states is lower than that of the current state. Otherwise, the algorithm resorts to a weighted A* search to select the next node to expand. In this latter case, the evaluation function used to rank the nodes is $f(S) = g(S) + w * h(S)$ where $g(S)$ is the length of the current partial plan in terms of number of steps, $h(S)$ is our estimated cost given by the heuristic function (e.g. $h_{AdjSum2M}$), and $w$ is the weight given to the heuristic function. $w$ is set to 5 based on our empirical experience.[2]

*Breaking Ties:* In case of a tie in selecting the pivot branch, i.e., more than one branch leads to a state with the lowest heuristic cost, we break the tie by choosing the action that

2. For the role of $w$ in Best-First search see (Korf, 1993).

supports subgoals that are harder to achieve. Here, the hardness of a literal $l$ is measured in terms of the level in the planning graph at which $l$ first appears. The standard rationale for this decision (c.f. Kambhampati & Sanchez, 2000) is that we want to fail faster by considering the most difficult subgoals first. We have an additional justification in our case, we also know that a subgoal with a higher level value requires more steps and actions for its achievement because it appeared later into the planning graph. So, by supporting it first, we may be able to achieve other easier subgoals along the way and thereby reduce the number of parallel steps in our partial plan.

**Fattening the Pivot Branch:** Next the procedure needs to decide which subset $O \subseteq A$ of the sibling actions of the pivot action $a_p$ will be used to fatten the pivot branch. The obvious first idea would be to fatten the pivot branch maximally by adding all pairwise independent actions found during that search stage. The problem with this idea is that it may add redundant and heuristically inferior actions to the branch, and satisfying their preconditions may lead to an increase of the number of parallel steps.

So, in order to avoid fattening the pivot branch with such irrelevant actions, before adding any action $a$ to $O$, we *require* that the heuristic cost of the state $S'$ that results from regressing $S$ over $O + a$ be strictly lower than that of $S$. This is in addition to the requirement that $a$ be pairwise independent with the current set of actions in $O$. This simple check also ensures that we do not add more than one action for supporting the same set of subgoals in $S$.

The overall procedure for fattening the pivot branch thus involves picking the next hardest subgoal $g$ in $S$ (with hardness measured in terms of the level of the subgoal in the planning graph), and finding the action $a_g \in A$ achieving $g$, which is pair-wise independent of all actions in $O$ and which, when added to $O$ and used to regress $S$, leads to a state $S'$ with the lowest heuristic cost, which in consequence should be lower than the cost of $S$. Once found, $a_g$ is then added to $O$, and the procedure is repeated. If there is more than one action that can be $a_g$, then we break ties by considering the degree of overlap between the preconditions of action $a_g$ and the set of actions currently in $O$. The degree of precondition overlap between $a$ and $O$ is defined as $|prec(a) \cap \{\cup_{o \in O} prec(o)\}|$. The action $a$ with higher degree of overlap is preferred as this will reduce the amount of additional work we will need to do to establish its preconditions. Notice that because of the fattening process, a search node may have multiple actions leading to it from its parent, and multiple actions leading from it to each of its children.

**Example:** Figure 5 illustrates the use of this node expansion procedure for a problem from the logistics domain (Bacchus, 2001). In this example we have four packages `pack1`, `pack2`, `pack3` and `pack4`. Our goal is to place the first three of them at `ASU` and the remaining one at `home`. There are two planes `airp1` and `airp2` to carry out the plans. The figure shows the first level of the search after $S$ has been regressed. It also shows the pivot action $a_p$ given by `unload(pack1,airp1,ASU)`, and a candidate set of pairwise independent actions with respect to $a_p$. Finally, we can see in Figure 6 the generation of the parallel branch. Notice that each node can be seen as a partial regressed plan. As described in the paragraphs above, only actions regressing to lower heuristic estimates are considered in $a_{par}$ to fatten the pivot branch. Notice that the action `unload(pack4,airp2,Home)` has been discarded because it leads to a state with higher cost, even though it is not inconsistent with the rest of the actions chosen to fatten the pivot branch. Furthermore, we can also see that we have preferred actions using the plane `airp1`, since they overlap more with the pivot action $a_p$.

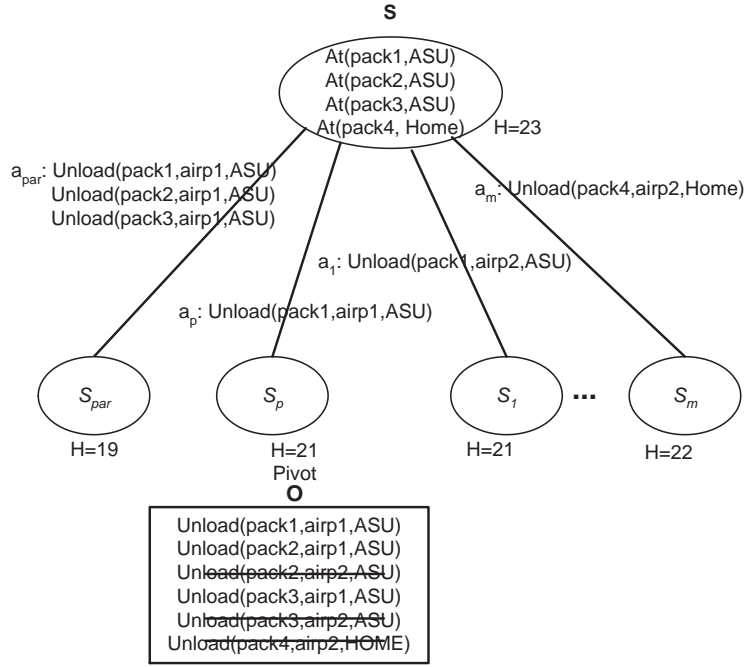

Figure 6: $S_{par}$ is the result of incrementally fattening the $Pivot$ branch with the pairwise independent actions in $O$

**Offsetting the Greediness of Fattening:** The fattening procedure is greedy, since it insists that the state resulting after fattening have a strictly better heuristic value. While useful in avoiding the addition of irrelevant actions to the plan, this procedure can also sometimes preclude actions that are ultimately relevant but were discarded because the heuristic is not perfect. These actions may then become part of the plan at later stages during search (i.e., earlier parts of the execution of the eventual solution plan; since we are searching in the space of plan suffixes). When this happens, the length of the parallel plan is likely to be greater, since more steps that may be needed to support the preconditions of such actions would be forced to come at even later stages of search (earlier parts of the plan). Had the action been allowed into the partial plan earlier in the search (i.e., closer to the end of the eventual solution plan), its preconditions could probably have been achieved in parallel to the other subgoals in the plan, thus improving the number of steps.

In order to offset this negative effect of greediness, $AltAlt^p$ re-arranges the partial plan to promote such actions higher up the search branch (i.e., later parts of the execution of the eventual solution plan). Specifically, before expanding a given node $S$, $AltAlt^p$ checks to see if any of the actions in $A_s$ leading to $S$ from its parent node (i.e., Figure 6 shows that $A_{par}$ leads to $S_{par}$) can be pushed up to higher levels in the search branch. This online

**pushUP**(S)
    $A_s \leftarrow$ *get actions leading to S*
    **forall** $a \in A_s$
        $x \leftarrow 0$
        $S_x \leftarrow$ *get parent node of S*
            /** Getting highest ancestor for each action
        **Loop**
            $A_x \leftarrow$ *get actions leading to $S_x$*
            **If** $(parallel(a, A_x))$
                $x \leftarrow x + 1$
                $S_x \leftarrow$ *get parent node of $S_{x-1}$*
            **Else**
                $a_j \leftarrow$ *get action conflicting with a from $A_x$*
                **If** *(Secondary Optimizations)*
                    *Remove a and $a_j$ from branch*
                    *Include $a_{new}$ if necessary*
                **Else**
                    $A_{x-1} \leftarrow A_{x-1} + a$
                $A_s \leftarrow A_s - a$
             *break*
        **End Loop**
            /**Adjusting the partial plan
        $S_x \leftarrow$ *get highest ancestor x in history*
        *createNewBranchFrom($S_x$)*
        **while** $x > 0$
            $S_{new} \leftarrow$ *regress $S_x$ with $A_{x-1}$*
            $S_x \leftarrow S_{new}$
            $x \leftarrow x - 1$
**END**;

Figure 7: *Pushup* Procedure

re-arrangement of the plan is done by the *Pushup* procedure, which is shown in Figure 7. The *Pushup* procedure is called each time before a node gets expanded, and it will try to compress the partial plan. For each of the actions $a \in A_s$ we find the highest ancestor node $S_x$ of $S$ in the search branch to which the action can be applied (i.e., it gives some literal in $S_x$ without deleting any other literals in $S_x$, and it is pairwise independent of all the actions $A_x$ currently leading out of $S_x$, in other words the condition $parallel(a, A_x)$ is satisfied). Once $S_x$ is found, $a$ is then removed from the set of actions $A_s$ leading to $S$ and introduced into the set of actions leading out of $S_x$ (to its child in the current search branch). Next, the states in the search branch below $S_x$ are adjusted to reflect this change. The adjustment involves recomputing the regressions of all the search nodes below $S_x$. At first glance, this might seem like a transformation of questionable utility since the preconditions of $a$ (and their regressions) just become part of the descendants of $S_x$, and this does not necessarily reduce the length of the plan. We however expect a length reduction because actions supporting the preconditions of $a$ will get "pushed up" eventually during later expansions.

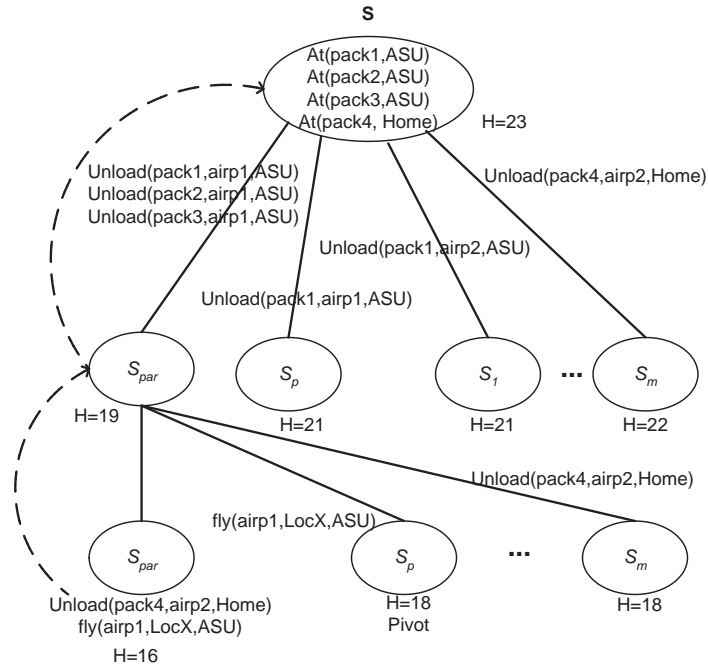

(a) Finding the highest ancestor node to which an action can be pushed up.

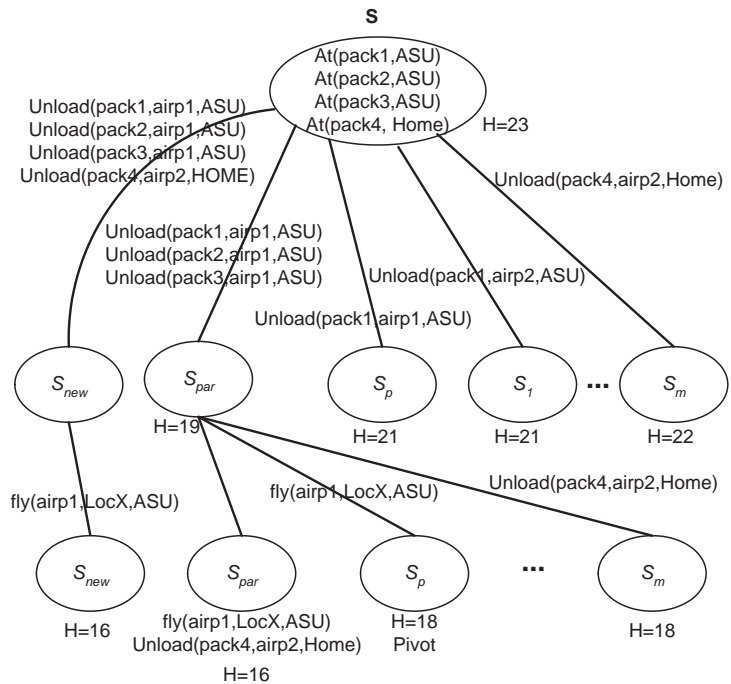

(b) The *Pushup* procedure generates a new search branch.

Figure 8: Rearranging of the Partial Plan

Rather than doctor the existing branch, in the current implementation, we just add a new branch below $S_x$ that reflects the changes made by the *Pushup* procedure.[3] The new branch then becomes the active search branch, and its leaf node is expanded next.

**Aggressive Variation of** *Pushup*: The *Pushup* procedure, as described above, is not expensive as it only affects the current search branch, and the only operations involved are recomputing the regressions in the branch. Of course, it is possible to be more aggressive in manipulating the search branch. For example, after applying an action $a$ to its ancestor $S_x$ the set of literals in the child state, say $S_{new}$ changes, and thus additional actions may become relevant for expanding $S_{new}$. In principle, we could re-expand $S_{new}$ in light of the new information. We decided not to go with the re-expansion option, as it typically does not seem to be worth the cost. In Section 4.3, we do compare our default version of *Pushup* procedure with a variant that re-expands all nodes in the search branch, and the results of those studies support our decision to avoid re-expansion. Finally, although we introduced the *Pushup* procedure as an add-on to the fattening step, it can also be used independent of the latter, in which case the net effect would be an incremental parallelization of a sequential plan.

**Example:** In Figure 8(a), we have two actions leading to the node $S_{par}$ (at depth two), these two actions are Unload(pack4,airp2,Home) and fly(airp1,LocX,ASU). So, before expanding $S_{par}$ we check if any of the two actions leading to it can be pushed up. While the second action is not pushable since it interacts with the actions in its ancestor node, the first one is. We find the highest ancestor in the partial plan that interacts with our pushable action. In our example the root node is such an ancestor. So, we insert our pushable action Unload(pack4,airp2,Home) directly below the root node. We then re-adjust the state $S_{par}$ to $S_{new}$ at depth 1, as shown in Figure 8(b), adding a new branch, and reflecting the changes in the states below. Notice that the action Unload(pack4,airp2,Home) was initially discarded by the greediness of the fattening procedure (see Figure 6), but we have offset this negative effect with our plan compression algorithm. We can see also that we have not re-expanded the state $S_{new}$ at depth 1, we have only made the adjustments to the partial plan using the actions already presented in the search trace.[4]

# 4. Evaluating the Performance of *AltAlt$^p$*

We implemented *AltAlt$^p$* on top of *AltAlt*. We have tested our implementation on a suite of problems that were used in the 2000 and 2002 AIPS competition (Bacchus, 2001; Long & Fox, 2002), as well as other benchmark problems (McDermott, 2000). Our experiments are broadly divided into three sets, each aimed at comparing the performance of *AltAlt$^p$* under different scenarios:

1. Comparing the performance of *AltAlt$^p$* to other planning systems capable of producing parallel plans.

---

3. Because of the way our data structures are set up, adding the new branch turns out to be a more robust option than manipulating the existing search branch.

4. Instead, the aggressive *Pushup* modification would expand $S_{new}$ at depth 1, generating similar states to those generated by the expansion of $S_{par}$ at the same depth.

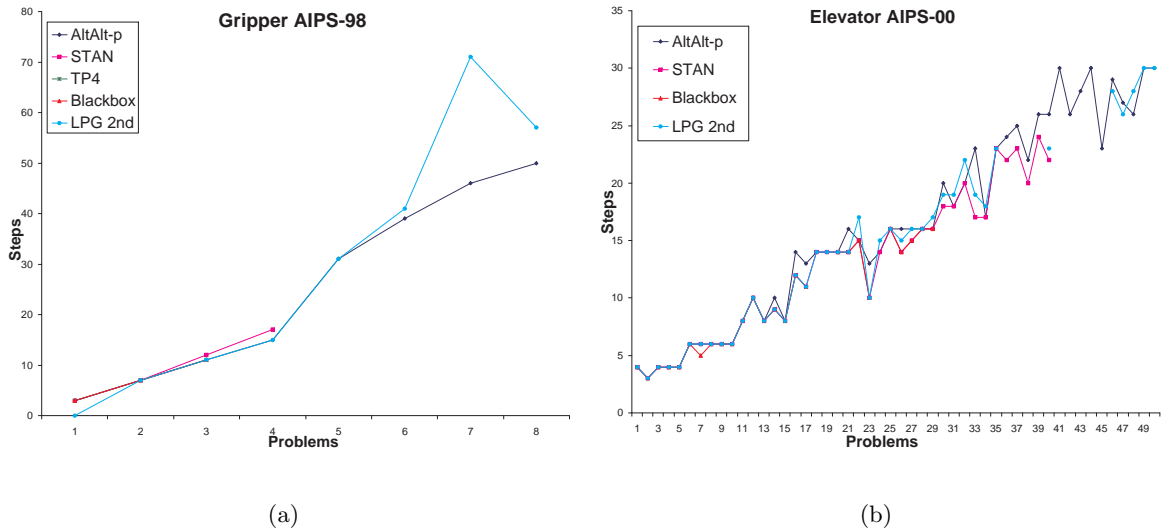

(a)                                                    (b)

Figure 9: Performance on the Gripper (AIPS-98) and the Elevator (AIPS-00) Domains.

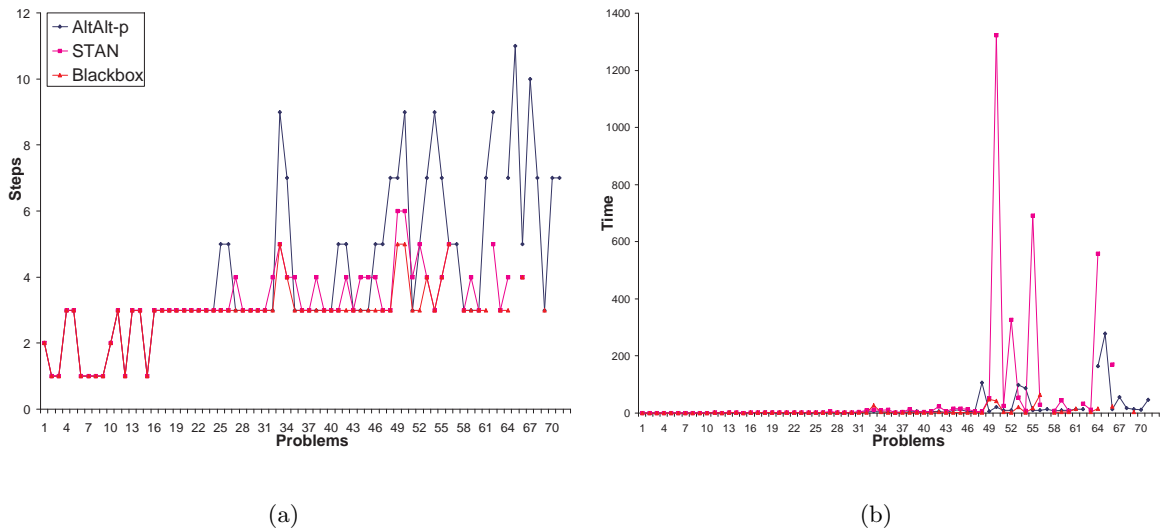

(a)                                                    (b)

Figure 10: Performance on the Schedule domain (AIPS-00)

2. Comparing our incremental parallelization technique to $AltAlt$ + Post-Processing.

3. Ablation studies to analyze the effect of the different parts of the $AltAlt^p$ approach on its overall performance.

Our experiments were all done on a Sun Blade-100 workstation, running SunOS 5.8 with 1GB RAM. Unless noted otherwise, $AltAlt^p$ was run with the $h_{adjsum2M}$ heuristic described in section 2 of this paper, and with a parallel planning graph grown until the first level where the top-level goals are present without being mutex. All times are in seconds.

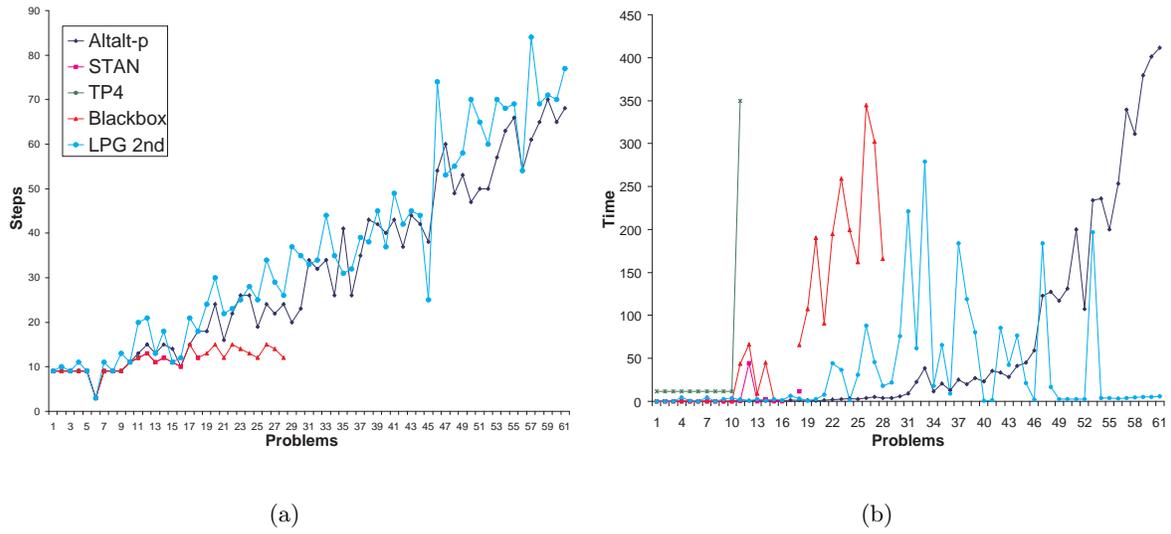

(a)                                                         (b)

Figure 11: Performance on the Logistics domain(AIPS-00)

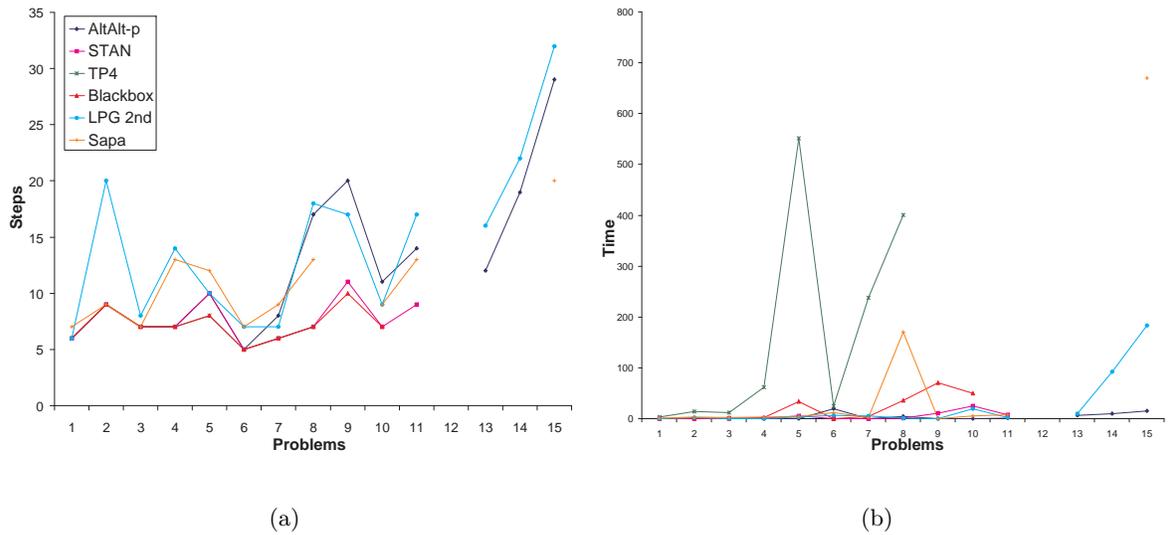

(a)                                                         (b)

Figure 12: Performance on the DriverLog domain(AIPS-02)

## 4.1 Comparing $AltAlt^p$ with Competing Approaches

In the first set of experiments we have compared the performance of our planner with the results obtained by running STAN (Long & Fox, 1999), Blackbox (Kautz & Selman, 1999), TP4 (Haslum & Geffner, 2001), LPG (Gerevini & Serina, 2002) and SAPA (Do & Kambhampati, 2001). Unless noted otherwise, every planner has been run with its default settings. Some of the planners could not be run in some domains due to parsing problems or memory allocation errors. In such cases, we just omit that planner from consideration for those particular domains.

### 4.1.1 PLANNERS USED IN THE COMPARISON STUDIES

STAN is a disjunctive planner, which is an optimized version of the Graphplan algorithm that reasons with invariants and symmetries to reduce the size of the search space. Blackbox is also based on the Graphplan algorithm but it works by converting planning problems specified in STRIPS (Fikes & Nilsson, 1971) notation into boolean satisfiability problems and solving it using a SAT solver (the version we used defaults to SATZ).[5] LPG (Gerevini & Serina, 2002) was judged the best performing planner at the 3rd International Planning Competition (Long & Fox, 2002), and it is a planner based on planning graphs and local search inspired by the Walksat approach. LPG was run with its default heuristics and settings. Since LPG employs an iterative improvement algorithm, the quality of the plans produced by it can be improved by running it for multiple iterations (thus increasing the running time). To make the comparisons meaningful, we decided to run LPG for two iterations (n=2), since beyond that, the running time of LPG was generally worse than that of $AltAlt^p$. Finally, we have also chosen two metric temporal planners, which are able to represent parallel plans because of their representation of time and durative actions. TP4 (Haslum & Geffner, 2001) is a temporal planner based on HSP*p (Haslum & Geffner, 2000), which is an optimal parallel state space planner with an IDA* search algorithm. The last planner in our list is SAPA (Do & Kambhampati, 2001). SAPA is a powerful domain-independent heuristic forward chaining planner for metric temporal domains that employs distance-based heuristics (Kambhampati & Sanchez, 2000) to control its search.

### 4.1.2 COMPARISON RESULTS IN DIFFERENT DOMAINS

We have run the planners in the Gripper domain from the International Planning and Scheduling competition from 1998 (McDermott, 2000), as well as three different domains (Logistics, Scheduling, and Elevator-miconic-strips) from 2000 (Bacchus, 2001), and three more from the 2002 competition (Long & Fox, 2002) - DriverLog, ZenoTravel, and Satellite. In cases where there were multiple versions of a domain, we used the "STRIPS Untyped" versions.[6]. We discuss the results of each of the domains below.

**Gripper:** In Figure 9(a), we compare the performance of $AltAlt^p$ on the Gripper domain (McDermott, 2000) to the rest of the planners excluding SAPA. The plot shows the results in terms of number of (parallel) steps. We can see that for even this simplistic domain, $AltAlt^p$ and LPG are the only planners capable of scaling up and generating parallel plans. None of the other approaches is able to solve more than four problems.[7] $AltAlt^p$ is able to scale up without any difficulty to problems involving 30 balls. Furthermore, $AltAlt^p$ returns better plans than LPG.

---

5. We have not chosen IPP (Koehler, 1999), which is also an optimized Graphplan planning system because results reported by Haslum and Geffner (2001) show that it is already less efficient than STAN.

6. Since SAPA does not read the STRIPS file format, we have run the SAPA planner on equivalent problems with unit-duration actions from Long and Fox (2002).

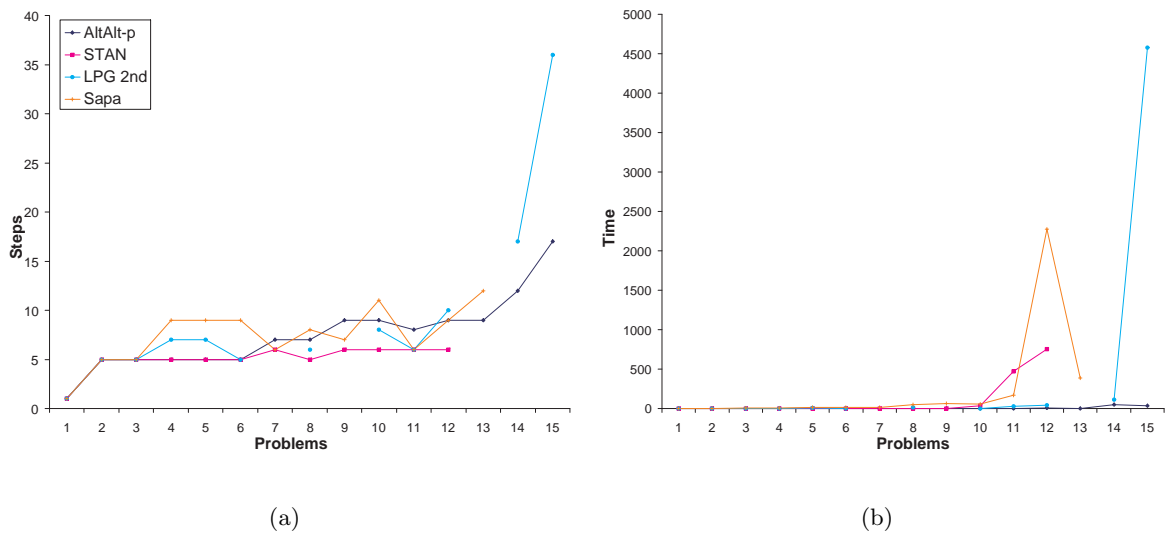

Figure 13: Performance on the ZenoTravel domain (AIPS-02)

**Elevator:** In Figure 9(b), we compare $AltAlt^p$ to STAN, Blackbox and LPG in the Elevator domain (Miconic Strips) (Bacchus, 2001).[8] $AltAlt^p$ approached the quality of the solutions produced by the optimal approaches (e.g. Blackbox and STAN). Notice that Blackbox can only solve around half of the problems solved by $AltAlt^p$ in this domain.

**Scheduling:** Results from the Scheduling domain are shown in Figure 10. Only Blackbox and STAN are considered for comparison.[9] $AltAlt^p$ seems to reasonably approximate the optimal parallel plans for many problems (around 50 of them), but does produce significantly suboptimal plans for some. However, it is again able to solve more problems than the other two approaches and in a fraction of the time.

**Logistics:** The plots corresponding to the Logistics domain from Bacchus (2001) are shown in Figure 11.[10] For some of the most difficult problems $AltAlt^p$ outputs lower quality solutions than the optimal approaches. However, only $AltAlt^p$ and LPG are able to scale up to more complex problems, and we can easily see that $AltAlt^p$ provides better quality solutions than LPG. $AltAlt^p$ also seems to be more efficient than any of the other approaches.

---

7. Although STAN is supposed to be able to generate optimal step-length plans, in a handful of cases it seems to have produced nonoptimal solutions for the Gripper Domain. We have no explanation for this behavior, but have informed the authors of the code.

8. we did not include the traces from TP4 because the pre-processor of the planner was not able to read the domain.

9. The TP4 pre-processor cannot read this domain, LPG runs out of memory, and SAPA has parsing problems.

10. Only SAPA is excluded due to parsing problems.

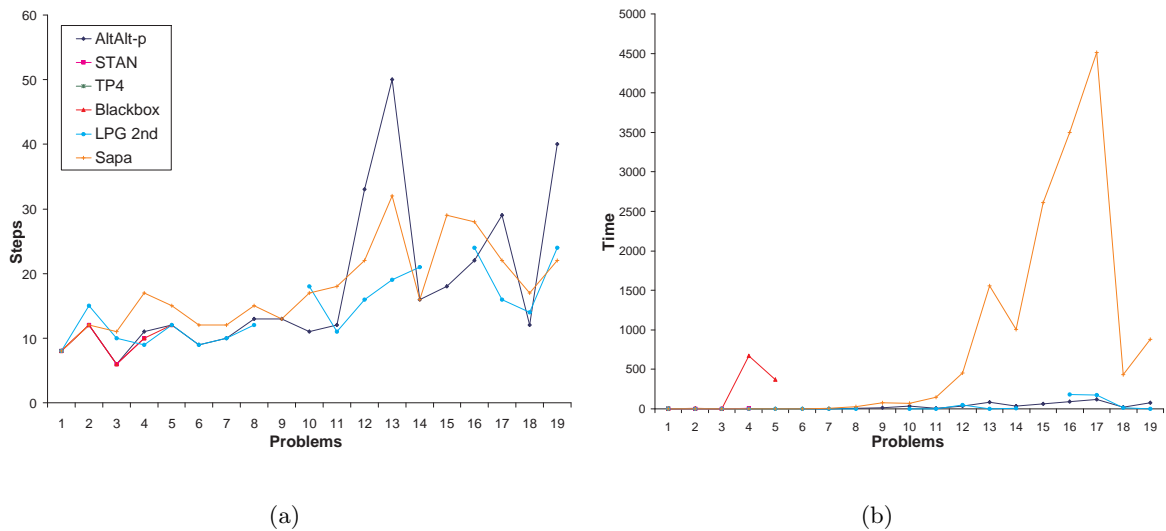

(a)                                                              (b)

Figure 14: Performance on the Satellite domain(AIPS-02)

The LPG solutions for problems 49 to 61 are obtained doing only one iteration, since LPG was not able to complete the second iteration in a reasonable amount of time. This explains the low time taken for LPG, and the lower quality of its solutions.

**DriverLog:** We see in Figure 12(a) that $AltAlt^p$ does reasonably well in terms of quality with respect to the other approaches in the DriverLog domain. Every planner is considered this time. $AltAlt^p$ is one of the two planners able to scale up. Figure 12(b) shows also that $AltAlt^p$ is more efficient than any of the other planners.

**Zeno-Travel:** Only $AltAlt^p$, SAPA, and LPG are able to solve most of the problems in this domain.[11] $AltAlt^p$ solves them very efficiently (Figure 13(b)) providing very good solution quality (Figure 13(a)) compared to the temporal metric planners.

**Satellite:** The results from the Satellite domain are shown in Figure 14. Although every planner is considered, only $AltAlt^p$, SAPA, and LPG can solve most of the problems. SAPA solves all problems but produces lower quality solutions for many of them. $AltAlt^p$ produces better solution quality than SAPA, and is also more efficient. However, $AltAlt^p$ produces lower quality solutions than LPG in four problems. LPG cannot solve one of the problems and produces lower quality solutions in 5 of them.

**Summary:** In summary, we note that $AltAlt^p$ is significantly superior in the elevator and gripper domains. It also performs very well in the DriverLog, ZenoTravel, and Satellite domains from the 2002 competition (Long & Fox, 2002). The performance of all planners is similar in the Schedule domain. In the Logistics domain, the quality of $AltAlt^p$ plans are second only to those of Blackbox for the problems that this optimal planner can solve. However, it scales up along with LPG to bigger size problems, returning very good step-length quality plans. TP4, the only other heuristic state search regression planner capable of producing parallel plans is not able to scale up in most of the domains. SAPA, a heuristic search progression planner, while competitive, is still outperformed by *AltAlt^p* in planning time and solution quality.

---

11. Blackbox and TP4 are not able to parse this domain.

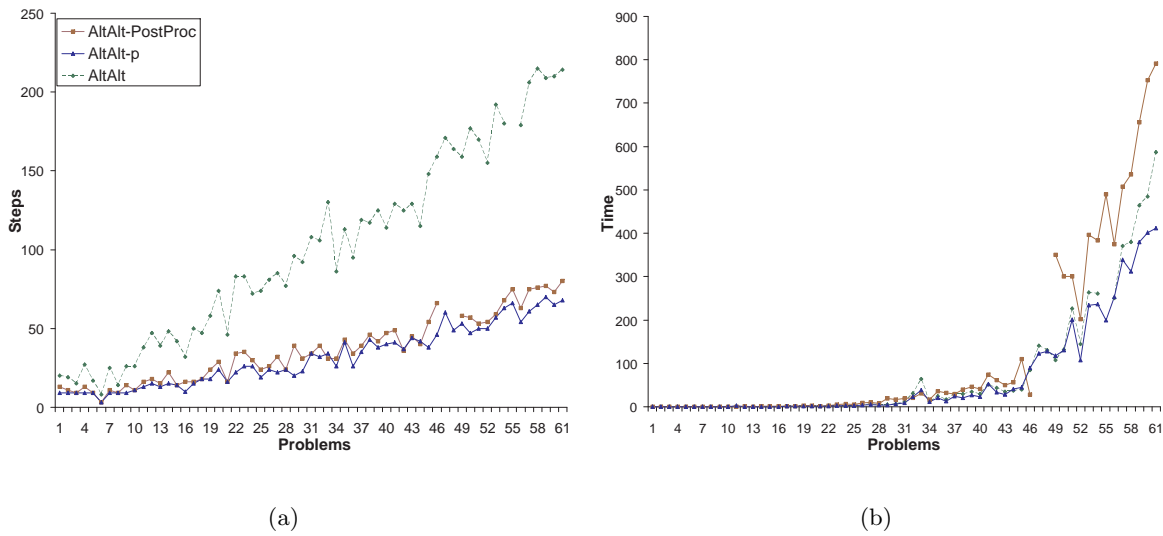

Figure 15: *AltAlt* and Post-Processing *vs.* *AltAlt^p* (Logistics domain)

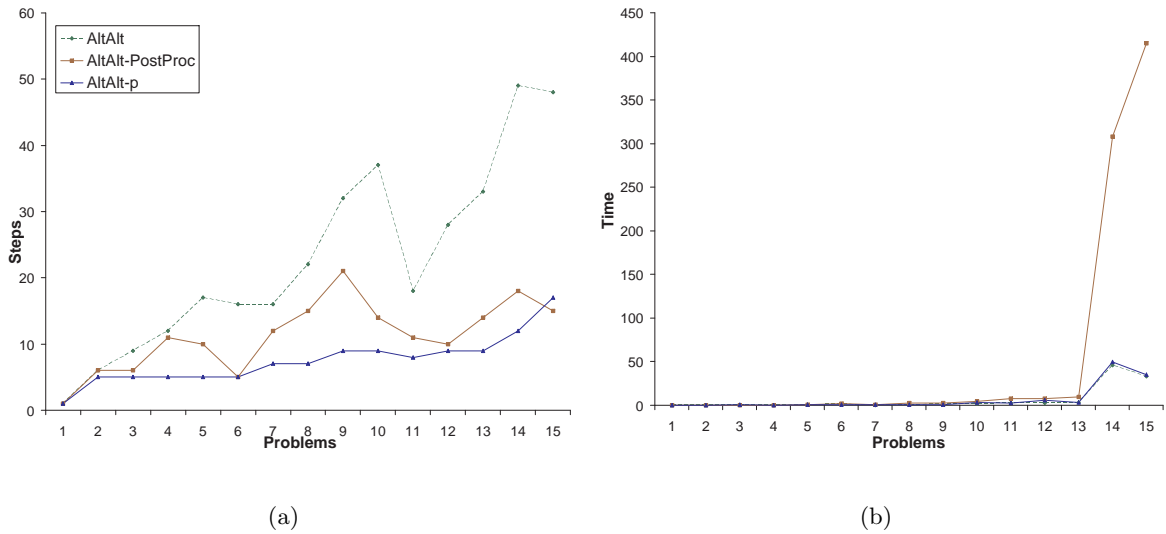

Figure 16: *AltAlt* and Post-Processing *vs.* *AltAlt^p* (Zenotravel domain)

SOLUTION: solution found (length = 9)
Time 1: load-truck(obj13,tru1,pos1) Level: 1
Time 1: load-truck(obj12,tru1,pos1) Level: 1
Time 1: load-truck(obj11,tru1,pos1) Level: 1
Time 2: drive-truck(tru1,pos1,apt1,cit1) Level: 1
Time 3: unload-truck(obj12,tru1,apt1) Level: 3
Time 3: fly-airplane(apn1,apt2,apt1) Level: 1
Time 3: unload-truck(obj11,tru1,apt1) Level: 3
Time 4: load-airplane(obj12,apn1,apt1) Level: 4
Time 4: load-airplane(obj11,apn1,apt1) Level: 4
Time 5: load-truck(obj21,tru2,pos2) Level: 1
Time 5: fly-airplane(apn1,apt1,apt2) Level: 2
Time 6: drive-truck(tru2,pos2,apt2,cit2) Level: 1
Time 6: unload-airplane(obj11,apn1,apt2) Level: 6
Time 7: load-truck(obj11,tru2,apt2) Level: 7
Time 7: unload-truck(obj21,tru2,apt2) Level: 3
Time 8: drive-truck(tru2,apt2,pos2,cit2) Level: 2
Time 9: unload-airplane(obj12,apn1,apt2) Level: 6
Time 9: unload-truck(obj13,tru1,apt1) Level: 3
Time 9: unload-truck(obj11,tru2,pos2) Level: 9
Total Number of actions in Plan: 19

POST PROCESSED PLAN ...
Time: 1 :  load-truck(obj13,tru1,pos1)
Time: 1 :  load-truck(obj12,tru1,pos1)
Time: 1 :  load-truck(obj11,tru1,pos1)
Time: 1 :  fly-airplane(apn1,apt2,apt1)
Time: 1 :  load-truck(obj21,tru2,pos2)
Time: 2 :  drive-truck(tru1,pos1,apt1,cit1)
Time: 2 :  drive-truck(tru2,pos2,apt2,cit2)
Time: 3 :  unload-truck(obj12,tru1,apt1)
Time: 3 :  unload-truck(obj11,tru1,apt1)
Time: 3 :  unload-truck(obj21,tru2,apt2)
Time: 3 :  unload-truck(obj13,tru1,apt1)
Time: 4 :  load-airplane(obj12,apn1,apt1)
Time: 4 :  load-airplane(obj11,apn1,apt1)
Time: 5 :  fly-airplane(apn1,apt1,apt2)
Time: 6 :  unload-airplane(obj11,apn1,apt2)
Time: 6 :  unload-airplane(obj12,apn1,apt2)
Time: 7 :  load-truck(obj11,tru2,apt2)
Time: 8 :  drive-truck(tru2,apt2,pos2,cit2)
Time: 9 :  unload-truck(obj11,tru2,pos2)
END OF POST PROCESSING: Actions= 19 Length: 9

(a) $AltAlt^p$ Solution

(b) $AltAlt^p$ plus Post-processing

Figure 17: Plots showing that $AltAlt^p$ solutions cannot be improved anymore by Post-processing.

## 4.2 Comparison to Post-Processing Approaches

As we mentioned earlier (see Section 1), one way of producing parallel plans that has been studied previously in the literature is to post-process sequential plans (Backstrom, 1998). To compare online parallelization to post-processing, we have implemented Backstrom (1998)'s "Minimal De-ordering Algorithm", and used it to post-process the sequential plans produced by $AltAlt$ (running with its default heuristic $h_{AdjSum2M}$ using a serial planning graph). In this section we will compare our online parallelization procedure to this offline method.

The first set of experiments is on the Logistics domain (Bacchus, 2001). The results are shown in Figure 15. As expected, the original $AltAlt$ has the longest plans since it allows only one action per time step. The plot shows that post-processing techniques do help in reducing the makespan of the plans generated by $AltAlt$. However, we also notice that $AltAlt^p$ outputs plans with better makespan than either $AltAlt$ or $AltAlt$ followed by post-processing. This shows that online parallelization is a better approach than post-processing sequential plans. Moreover, the plot in Figure 15(b) shows that the time taken by $AltAlt^p$ is largely comparable to that taken by the other two approaches. In fact, there is not much additional cost overhead in our procedure.

Figure 16 repeats these experiments in the ZenoTravel domain (Long & Fox, 2002). Once again, we see that $AltAlt^p$ produces better makespan than post-processing the sequential plans of $AltAlt$. Notice that this time, $AltAlt$ plus post-processing is clearly less efficient than either of the other two approaches. In summary, the results of this section demonstrate that $AltAlt^p$ is superior to $AltAlt$ plus post-processing.

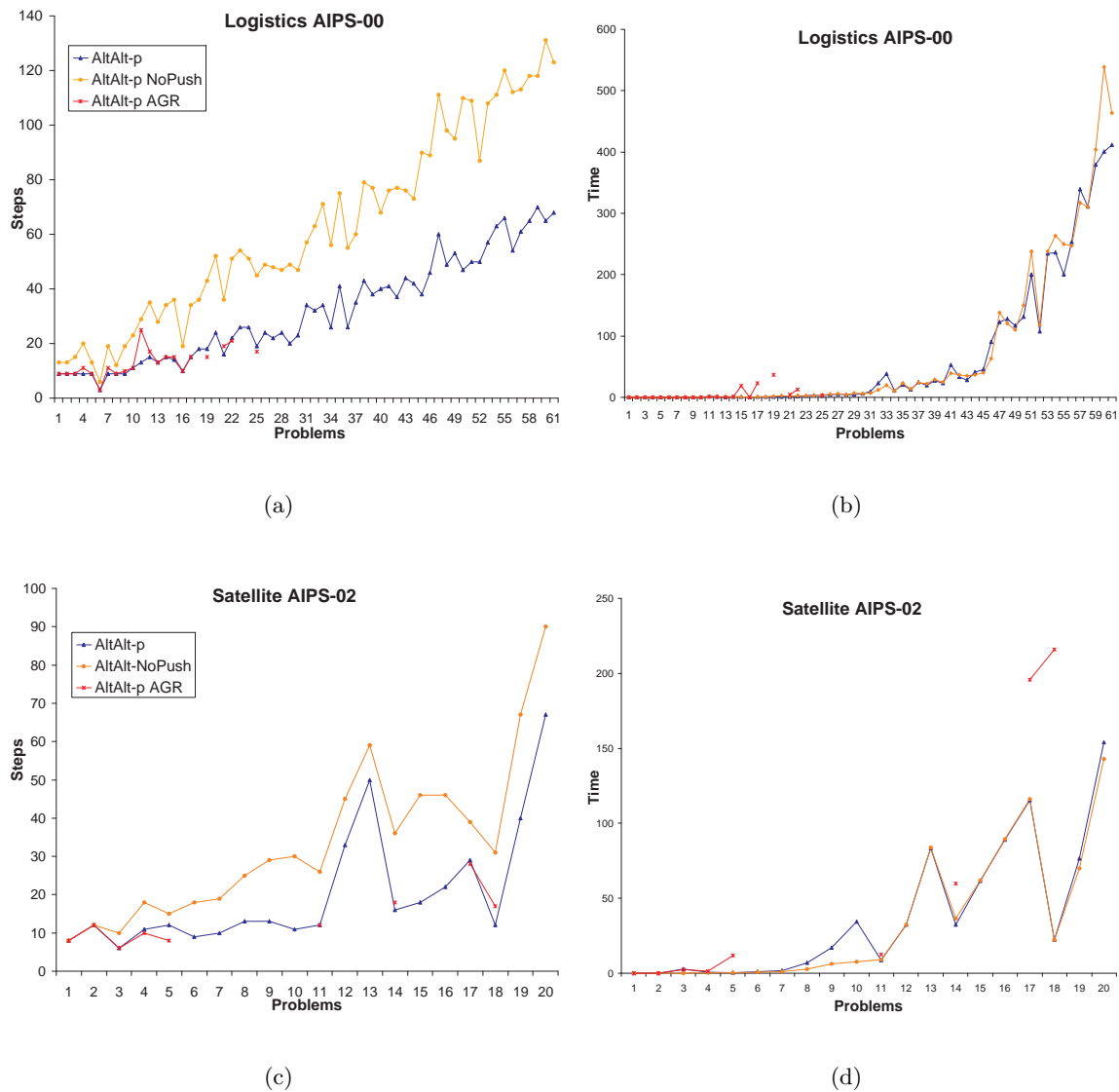

(a)

(b)

(c)

(d)

Figure 18: Analyzing the effect of the *Pushup* procedure

One might wonder if the plans generated by $AltAlt^p$ can also benefit from the post-processing phase. We have investigated this issue and found that the specific post-processing routines that we used do not produce any further improvements. The main reason for this behavior is that the *Pushup* procedure already tries to exploit any opportunity for shortening the plan length by promoting actions up in the partial plan. As an illustrative example, we show, in Figure 17, the parallel plan output by $AltAlt^p$ for a problem from the logistics domain (`logistics-4-1` from Bacchus, 2001), and the result of post-processing

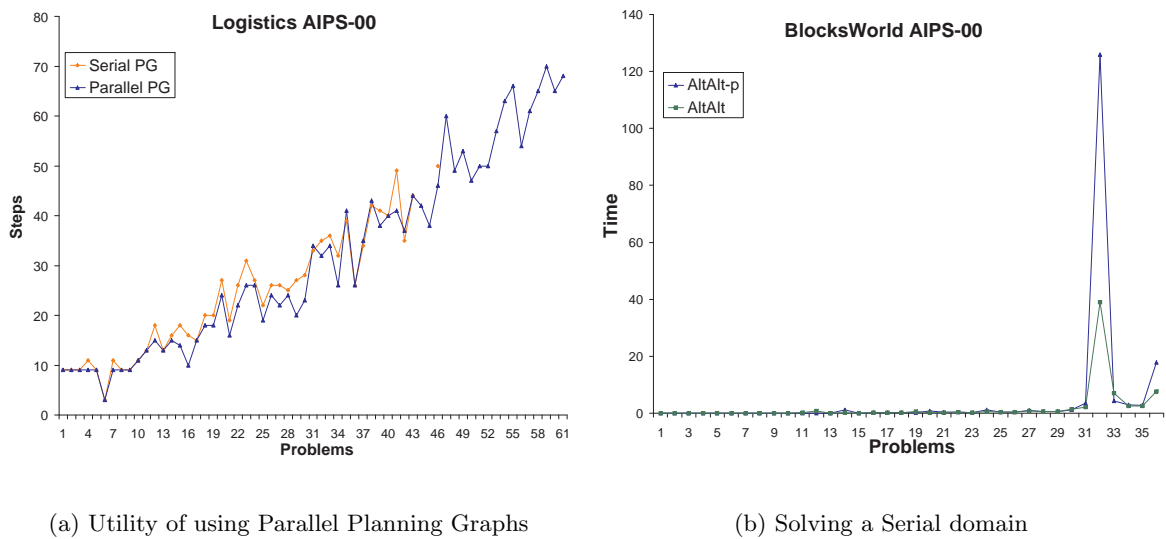

(a) Utility of using Parallel Planning Graphs          (b) Solving a Serial domain

Figure 19: Plots showing the utility of using parallel planning graphs in computing the heuristics, and characterizing the overhead incurred by $AltAlt^p$ in serial domains.

this solution. Although the two solutions differ in terms of step contents, we notice that they have the same step length. The difference in step contents can be explained by the fact that the de-ordering algorithm relaxes the ordering relations in the plan, allowing for some actions to come earlier, while *Pushup* always moves actions towards the end of the plan. We have run more comprehensive studies in three different domains (Logistics, Satellite and Zenotravel), and found that in no case is the step length of a plan produced by $AltAlt^p$ improved by the post-processing routine (we omit the comparison plots since they essentially show the curves corresponding to $AltAlt^p$ and $AltAlt^p$ with post-processing *coincident*).[12]

## 4.3 Ablation Studies

This section attempts to analyze the impact of the different parts of $AltAlt^p$ on its performance.

**Utility of the *Pushup* Procedure:** Figure 18 shows the effects of running $AltAlt^p$ with and without the *Pushup* procedure (but with the fattening procedure), as well as running it with a more aggressive version of *Pushup*, which as described in Section 3, re-expands all the nodes in the search branch, after an action has been pushed up. We can see that running $AltAlt^p$ with the *Pushup* and fattening procedures is better than just the latter. Comparison of results in Figure 15(a) and Figure 18(a) shows that even just the fattening procedure performs better than the original *AltAlt*. In Figure 18(b) we can see that although the *Pushup* procedure does not add much overhead, the aggressive version of *Pushup* does get quite expensive. We also notice that only around 20 problems are solved within time limits with aggressive *Pushup*. The plots in Figure 18(c) and Figure 18(d) show the results of the same experiments in the Satellite domain. We see that the situation is quite similar in this domain. We can conclude then that the *Pushup* procedure, used to offset the greediness of the algorithm, achieves its purpose.

---

12. We also verified this with at least one problem in all the other domains.

**Utility of basing heuristics on Parallel Planning Graphs:** We can see in Figure 19(a) that using parallel planning graph as the basis for deriving heuristic estimates in *AltAlt$^p$* is a winning idea. The serial planning graph overestimates the heuristic values in terms of steps, producing somewhat longer parallel solutions. The fact that the version using serial planning graph runs out of time in many problems also demonstrates that the running times are also improved by the use of parallel planning graphs.

**Comparison to *AltAlt*:** One final concern would be how much of an extra computational hit is taken by the *AltAlt$^p$* algorithm in serial domains (e.g. Blocks-World from Bacchus, 2001). We expect it to be negligible and to confirm our intuitions, we ran *AltAlt$^p$* on a set of problems from the sequential Blocks-World domain. We see from the plot 19(b) that the time performance between *AltAlt* and *AltAlt$^p$* are equivalent for almost all of the problems.

## 5. Related work

The idea of partial exploration of parallelizable sets of actions is not new (Kabanza, 1997; Godefroid & Kabanza, 1991; Do & Kambhampati, 2001). It has been studied in the area of concurrent and reactive planning, where one of the main goals is to approximate optimal parallelism. However, most of the research there has been focused on forward chaining planners (Kabanza, 1997), where the state of the world is completely known. It has been implied that backward-search methods are not suitable for this kind of analysis (Godefroid & Kabanza, 1991) because the search nodes correspond to partial states. We have shown that backward-search methods can also be used to approximate parallel plans in the context of classical planning.

Optimization of plans according to different criteria (e.g. execution time, quality, etc) has also been done as a post-processing step. The post-processing computation of a given plan to maximize its parallelism has been discussed by Backstrom (1998). Reordering and de-ordering techniques are used to maximize the parallelism of the plan. In de-ordering techniques ordering relations can only be removed, not added. In reordering, arbitrary modifications to the plan are allowed. In the general case this problem is NP-Hard and it is difficult to approximate (Backstrom, 1998). Furthermore, as discussed in Section 1 and 4, post-processing techniques are just concerned with modifying the order between the existing actions of a given plan. Our approach not only considers modifying such orderings but also inserting new actions online which can minimize the possible number of parallel steps of the overall problem.

We have already discussed Graphplan based planners (Long & Fox, 1999; Kautz & Selman, 1999), which return optimal plans based on the number of time steps. Graphplan uses IDA* to include the greatest number of parallel actions at each time step of the search. However, this iterative procedure is very time consuming and it does not provide any guarantee on the number of actions in its final plans. There have been a few attempts to minimize the number of actions in these planners (Huang, Selman, & Kautz, 1999) by using some domain control knowledge based on the generation of rules for each specific planning domain. The Graphplan algorithm tries to maximize its parallelism by satisfying most of the subgoals at each time step, if the search fails then it backtracks and reduces the set of parallel actions being considered one level before. *AltAlt$^p$* does the opposite, it tries to guess initial parallel nodes given the heuristics, and iteratively adds more actions to these nodes as possible with the *Pushup* procedure later during search.

More recently, there has been some work on generalizing forward state search to handle action concurrency in metric temporal domains. Of particular relevance to this work are the Temporal TLPlan (Bacchus & Ady, 2001) and SAPA (Do & Kambhampati, 2001). Both these planners are designed specifically for handling metric temporal domains, and use similar search strategies. The main difference between them being that Temporal TLPlan depends on hand-coded search control knowledge to guide its search, while SAPA (like *AltAlt$^p$*) uses heuristics derived from (temporal) planning graphs. As such, both these planners can be co-opted to produce parallel plans in classical domains. Both these planners do a forward chaining search, and like *AltAlt$^p$*, both of them achieve concurrency incrementally, without projecting sets of actions, in the following way. Normal forward search planners start with the initial state $S_0$, corresponding to time $t_0$, consider all actions that apply to $S_0$, and choose one, say $a_1$ apply it to $S_0$, getting $S_1$. They simultaneously progress the "system clock" from $t_0$ to $t_1$. In order to allow for concurrency, the planners by Bacchus and Ady (2001), and Do and Kambhampati (2001) essentially decouple the "action application" and "clock progression." At every point in the search, there is a non-deterministic choice - between progressing the clock, or applying (additional) actions at the current time point. From the point of view of these planners, *AltAlt$^p$* can be seen as providing heuristic guidance for this non-deterministic choice (modulo the difference that *AltAlt$^p$* does regression search). The results of empirical comparisons between *AltAlt$^p$* and SAPA, which show that *AltAlt$^p$* outperforms SAPA, suggest that the heuristic strategies employed in *AltAlt$^p$* including the incremental fattening, and the pushup procedure, can be gainfully adapted to these planners to increase the concurrency in the solution plans. Finally, HSP*, and TP4, its extension to temporal domains, are both heuristic state search planners using regression that are capable of producing parallel plans (Haslum & Geffner, 2000). TP4 can be seen as the regression version of the approach used in SAPA and temporal TLPlan. Our experiments however demonstrate that neither of these planners scales well in comparison to *AltAlt$^p$*.

The *Pushup* procedure can be seen as a plan compression procedure. As such, it is similar to other plan compression procedures such as "double-back optimization" (Crawford, 1996). One difference is that double-back is used in the context of a local search, while *Pushup* is used in the context of a systematic search. Double-back could be also applied to any finished plan or schedule, but as any other post-processing approach its outcome would depend highly on the plan given as input.

## 6. Concluding Remarks

Motivated by the acknowledged inability of heuristic search planners to generate parallel plans, we have developed and presented an approach to generate parallel plans in the context of *AltAlt*, a heuristic state space planner. This is a challenging problem because of the exponential branching factor incurred by naive methods. Our approach tries to avoid the branching factor blow up by greedy and online parallelization of the evolving partial plans. A plan compression procedure called *Pushup* is used to offset the ill effects of the greedy search. Our empirical results show that in comparison to other planners capable of producing parallel plans, *AltAlt$^p$* is able to provide reasonable quality parallel plans in a fraction of the time of competing approaches. Our approach also seems to provide better quality plans than can be achieved by post-processing sequential plans. Our results show that *AltAlt$^p$* provides an attractive tradeoff between quality and efficiency in the generation of parallel plans. In the future, we plan to adapt the *AltAlt$^p$* approach to metric temporal domains, where the need for concurrency is more pressing. One idea is to adapt some of the sources of strength in *AltAlt$^p$* to SAPA, a metric temporal planner being developed in our group (Do & Kambhampati, 2001).

## Acknowledgments

We thank Minh B. Do and XuanLong Nguyen for helpful discussions and feedback. We also thank David Smith and the JAIR reviewers for many constructive comments. This research is supported in part by the NASA grants NAG2-1461 and NCC-1225, and the NSF grant IRI-9801676.